\documentclass[runningheads]{llncs}
\usepackage[T1]{fontenc}
\usepackage{graphicx,verbatim}
\usepackage{booktabs}
\usepackage{amsmath,amssymb}
\begin{document}
\title{DARIL: When Imitation Learning outperforms Reinforcement Learning in Surgical Action Planning}
\titlerunning{When Imitation Learning outperforms RL in Surgical Planning}

% \author{Anonymized Authors}
% \authorrunning{Anonymized Authors et al.}
% \institute{Anonymized Affiliations \\
%     \email{email@anonymized.com}}
\author{Maxence Boels \and
Harry Robertshaw \and
Thomas C Booth \and
Prokar Dasgupta \and
Alejandro Granados \and
Sebastien Ourselin}
\authorrunning{M. Boels et al.}
\institute{Surgical and Interventional Engineering, King's College London, London, UK \\
    \email{maxence.boels@kcl.ac.uk}}

\maketitle
\begin{abstract}
Surgical action planning requires predicting future instrument-verb-target triplets for real-time assistance. While teleoperated robotic surgery provides natural expert demonstrations for imitation learning (IL), reinforcement learning (RL) could potentially discover superior strategies through exploration. We present the first comprehensive comparison of IL versus RL for surgical action planning on CholecT50. Our Dual-task Autoregressive Imitation Learning (DARIL) baseline achieves 34.6\% action triplet recognition mAP and 33.6\% next frame prediction mAP with smooth planning degradation to 29.2\% at 10-second horizons. We evaluated three RL variants: world model-based RL, direct video RL, and inverse RL enhancement. Surprisingly, all RL approaches underperformed DARIL—world model RL dropped to 3.1\% mAP at 10s while direct video RL achieved only 15.9\%. Our analysis reveals that distribution matching on expert-annotated test sets systematically favors IL over potentially valid RL policies that differ from training demonstrations. This challenges assumptions about RL superiority in sequential decision making and provides crucial insights for surgical AI development.

\keywords{Surgical Action Planning \and Imitation Learning \and Reinforcement Learning \and Temporal Planning \and Surgical AI}
\end{abstract}

\section{Introduction}

Surgical action planning, predicting future instrument-verb-target relationships in surgical videos, represents a critical component for real-time surgical assistance systems. Accurate future prediction is essential for enabling proactive surgical guidance, reducing surgeon cognitive load, and facilitating autonomous robotic assistance in complex procedures. While prior work has predominantly focused on recognition tasks~\cite{nwoye2020recognition,nwoye2022rendezvous,nwoye2023cholectriplet2021}, prospective action planning presents unique challenges requiring multi-horizon prediction capabilities under safety-critical constraints.

The fundamental question for surgical AI systems is the optimal learning paradigm: should systems learn through imitation of expert demonstrations (IL) or through trial-and-error optimization via reinforcement learning (RL)~\cite{boels2025surgical}? Teleoperated robotic surgery provides natural access to expert demonstrations, making IL attractive. However, RL could theoretically discover strategies beyond expert-level performance through exploration. Recent work in endovascular surgery has demonstrated RL's potential for autonomous navigation tasks, with successful applications in mechanical thrombectomy achieving high success rates while maintaining safety constraints~\cite{robertshaw2024autonomous,robertshaw2025reinforcement}. More broadly, RL world models in particular have shown that single configurations with no hyperparameter tuning can outperform specialized methods across diverse benchmark tasks, complete farsighted tasks such as collecting diamonds in Minecraft without human data or curricula, and capture expectations of future events during autonomous driving~\cite{hafner2023mastering,hansen2023td,hu2023gaia}.

Recent advances in surgical gesture prediction~\cite{shi2022recognition,weerasinghe2024multimodal} and vision transformers for surgical analysis~\cite{kiyasseh2023vision,liu2023skit} have shown promise, while long-term workflow prediction~\cite{boels2025swag} demonstrates the potential for anticipatory systems. Yet the comparative effectiveness of IL versus RL for surgical action planning remains unexplored.

This work addresses this gap through the first systematic comparison of IL and RL approaches for surgical action planning. Using the CholecT50 dataset~\cite{nwoye2022rendezvous}, we evaluate recognition accuracy and planning capability across multiple time horizons.

\textbf{Contributions}: (1) \textit{First systematic IL vs RL comparison}: Comprehensive evaluation addressing fundamental methodological questions in surgical AI. (2) \textit{Surprising negative results}: RL methods consistently underperform IL with world model RL dropping to 3.1\% mAP vs 29.2\% for our Dual-task Autoregressive Imitation Learning (DARIL) at 10s planning. (3) \textit{Novel DARIL architecture}: Dual-task autoregressive approach maintaining robust temporal consistency (34.6\% action triplet recognition degrading smoothly to 29.2\% at 10s). (4) \textit{Evaluation bias insights}: Analysis of how expert demonstration alignment systematically favors IL over valid RL policies.

\section{Methods}

\subsection{Problem Formulation}

Given surgical video frames $\{f_1, f_2, ..., f_t\}$, we predict future action triplets $\{a_{t+1}, a_{t+2}, ..., a_{t+H}\}$ where $H$ represents the planning horizon. Each triplet $a_i = (I_i, V_i, T_i)$ consists of instrument, verb, and target components from predefined vocabularies. Importantly, each frame can contain multiple simultaneous actions (0-3 per frame) due to multiple robotic arms with different instruments, making the problem highly sparse with 100 distinct action classes total. We evaluate both current action recognition and future action prediction with single-step ($H=1$) next frame prediction and multi-step prediction ($H>1$) for planning assessment across longer horizons.

\subsection{Dataset}

We use CholecT50~\cite{nwoye2022rendezvous}, containing 50 laparoscopic cholecystectomy videos with frame-level annotations for 100 distinct triplet classes. Following the standard evaluation protocol~\cite{nwoye2022data}, we use videos [2,6,14,23,25,50,51,66,79,111] for testing and the remaining 40 videos for training. The training set contains 78,968 frames while the test set contains 21,895 frames, representing expert-level surgical demonstrations at 1 FPS sampling.

\subsection{Dual-task Autoregressive Imitation Learning (DARIL)}

Our IL baseline follows an offline Behavior Cloning (BC) approach, using supervised learning on expert demonstrations to model surgical action prediction as causal sequence generation, combining frame-level processing with autoregressive action generation:

\begin{equation}
p(a_{t+1}|f_{t-w+1:t}) = \text{GPT-2}(\text{FrameEmb}(f_{t-w+1:t}))
\end{equation}

where $w=20$ represents the context window size.

\textbf{Architecture}: The model processes 1024-dimensional Swin transformer features~\cite{liu2021swin} through: (1) BiLSTM encoder for temporal current action recognition, (2) GPT-2 decoder~\cite{radford2019language} for causal future action generation using a context window of $w=20$ frame embeddings as input, and (3) separate prediction heads for the combined <instrument, verb, and target> class and surgical phase components.

\textbf{Training}: Dual-task optimization combines current action recognition and next action prediction losses, with additional auxiliary losses:
\begin{equation}
\mathcal{L} = \mathcal{L}_{\text{current}} + \mathcal{L}_{\text{next}} + \mathcal{L}_{\text{embed}} + \mathcal{L}_{\text{phase}}
\end{equation}
where $\mathcal{L}_{\text{current}} = -\sum_{t} \log p(a_t|f_{t-w+1:t})$ and $\mathcal{L}_{\text{next}} = -\sum_{t} \log p(a_{t+1}|f_{t-w+1:t})$ represent cross-entropy losses for direct prediction of the 100 action classes, $\mathcal{L}_{\text{embed}} = \sum_{t} ||e_{t+1} - \hat{e}_{t+1}||^2$ is the MSE loss for next frame embedding prediction, and $\mathcal{L}_{\text{phase}}$ is the phase recognition loss.

\subsection{Reinforcement Learning Approaches}

For reproducibility, we detail our RL problem formulation: we define states as sequences of frame embeddings, actions as predicted triplets, and design reward functions based on expert demonstration matching using cosine similarity between predicted and ground truth action sequences.

\textbf{Latent World Model + RL}: Following Dreamer~\cite{hafner2023mastering}, the current state-of-the-art for image-based RL tasks, we learn an action-conditioned world model predicting future states and rewards: $p(s_{t+1}, r_t|s_t, a_t)$. PPO~\cite{schulman2017proximal} trains policies in the learned latent environment with rewards designed for expert demonstration matching.

\textbf{Direct Video RL}: Model-free RL applied directly to video sequences using expert demonstration matching rewards. We evaluate PPO~\cite{schulman2017proximal} and A2C~\cite{mnih2016asynchronous}. These algorithms were selected as representative model-free methods with proven stability for continuous control tasks. We performed careful hyperparameter optimization, treating frame sequences as states and predicted triplets as actions.

\textbf{Inverse RL Enhancement}: Maximum Entropy IRL~\cite{ziebart2008maximum} learns reward functions from expert trajectories. We generate negative examples by sampling actions deviating from expert demonstrations, then use learned rewards to guide policy optimization while maintaining IL baseline performance.

\subsection{Evaluation Framework}

\textbf{Recognition Evaluation}: Standard mAP computation on current and next action predictions using IVT metrics~\cite{nwoye2022data}.

\textbf{Planning Evaluation}: Multi-horizon assessment across 1s, 2s, 3s, 5s, 10s, and 20s using mAP degradation analysis.

\textbf{Component Analysis}: Individual performance analysis for instruments, verbs, targets, and their combinations (IV, IT, IVT) to understand degradation patterns.

\section{Results}

\subsection{Main Comparative Results}

Table~\ref{tab:main_results} presents our experimental findings. DARIL achieves 34.6\% action triplet recognition mAP and 33.6\% next frame prediction mAP, consistently outperforming all RL variants across planning horizons. The performance gaps are substantial—world model RL drops to 3.1\% at 10s while DARIL maintains 29.2\%.

\begin{table}[h]
\centering
\caption{IL vs RL Performance Comparison. All values are IVT mAP (\%). Current refers to action triplet recognition, 1s/5s/10s refer to next frame prediction at different horizons.}
\label{tab:main_results}
\begin{tabular}{lcccc}
\toprule
\textbf{Method} & \textbf{Current} & \textbf{1s} & \textbf{5s} & \textbf{10s} \\
\midrule
DARIL (Ours) & \textbf{34.6} & \textbf{33.6} & \textbf{31.2} & \textbf{29.2} \\
DARIL + IRL & 33.1 & 32.1 & 29.6 & 28.1 \\
DARIL + Direct Video RL & 33.2 & 22.6 & 19.3 & 15.9 \\
Latent World Model + RL & 33.1 & 14.0 & 9.1 & 3.1 \\
\bottomrule
\end{tabular}
\end{table}

\subsection{Component-wise Analysis}

Table~\ref{tab:component_analysis} shows DARIL's component-wise performance. Instruments demonstrate highest stability (91.4\% to 88.2\%), while targets show more variability (52.7\% to 52.5\%). The IVT combination reflects multiplicative effects of constituent components.

\begin{table}[h]
\centering
\caption{DARIL Component-wise Performance Analysis. Current refers to action triplet recognition, Next refers to next frame prediction.}
\label{tab:component_analysis}
\begin{tabular}{lcccccc}
\toprule
\textbf{Component} & \textbf{Current} & \textbf{Next} & \textbf{Component} & \textbf{Current} & \textbf{Next} \\
\midrule
Instrument (I) & 91.4 & 88.2 & Instrument-Verb (IV) & 42.9 & 38.8 \\
Verb (V) & 69.4 & 68.1 & Instrument-Target (IT) & 43.5 & 43.6 \\
Target (T) & 52.7 & 52.5 & IVT & 34.6 & 33.6 \\
\bottomrule
\end{tabular}
\end{table}

\subsection{Planning Performance Analysis}

\begin{figure}[h]
\centering
\includegraphics[width=\textwidth]{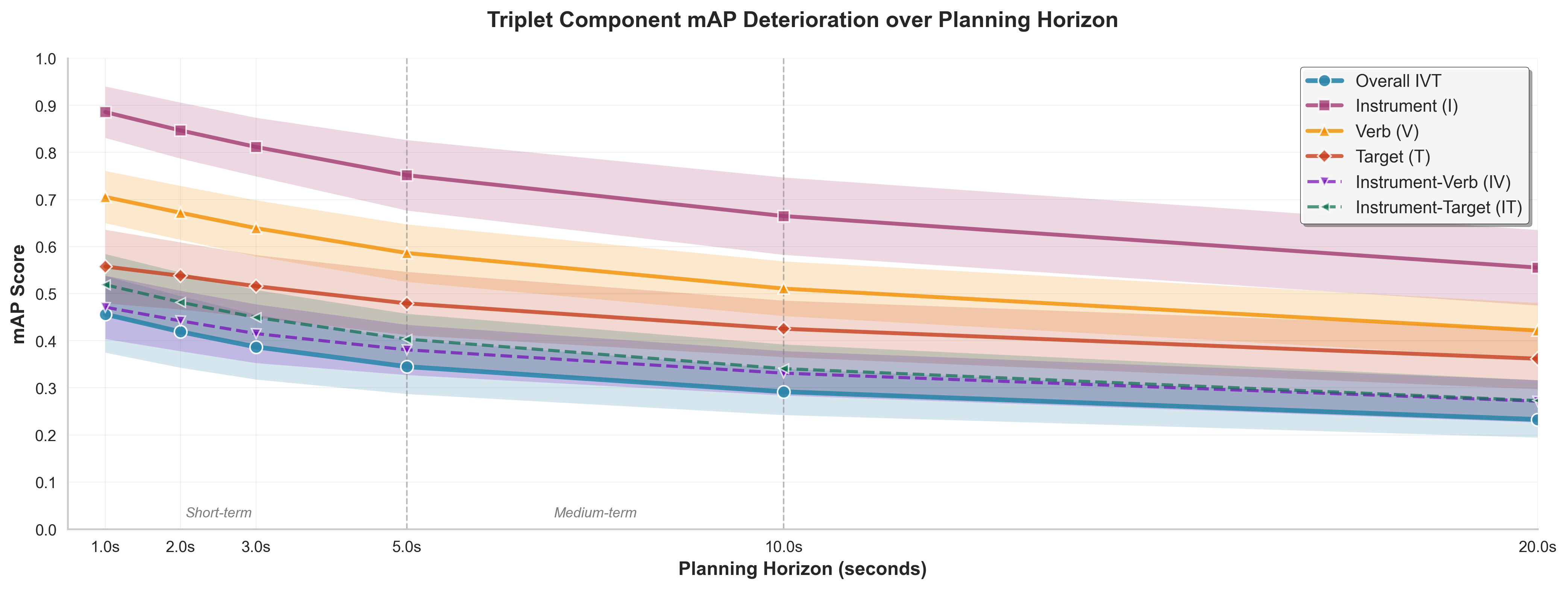}
\caption{DARIL planning performance across time horizons. The model maintains stable performance across different IVT mAP score components with graceful degradation over longer planning horizons. Error bars indicate 95\% confidence intervals.}
\label{fig:planning_analysis}
\end{figure}

Figure~\ref{fig:planning_analysis} demonstrates DARIL's smooth degradation across horizons from 33.6\% at 1s to 29.2\% at 10s (13.1\% relative decrease). This demonstrates the model's robust temporal consistency across different planning horizons.

\subsection{Qualitative Analysis}

\begin{figure}[h]
\centering
\includegraphics[width=\textwidth]{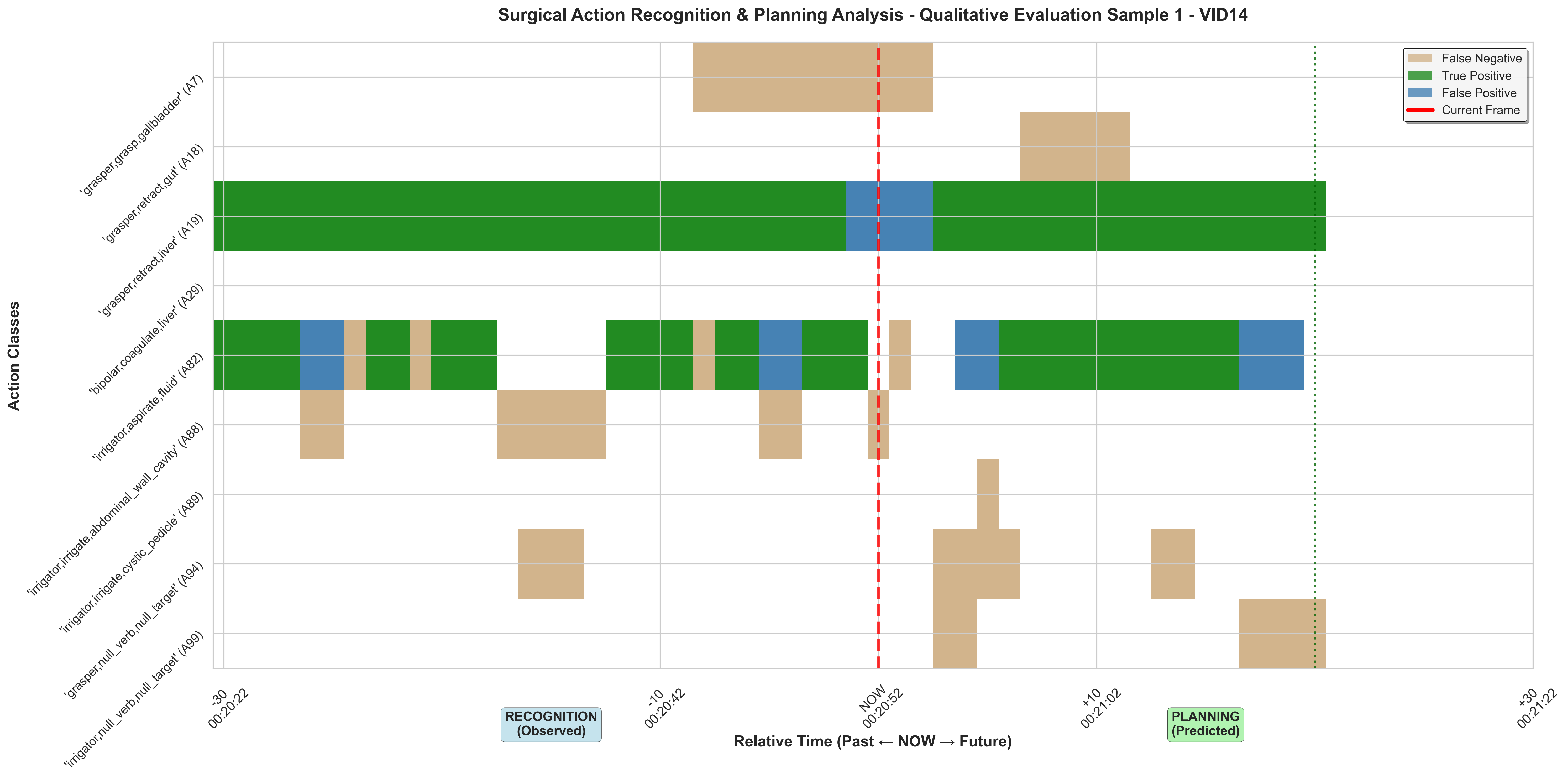}
\caption{Qualitative evaluation showing recognition (past) and planning (future) performance. Green indicates true positives, blue shows false positives, beige represents false negatives. The model demonstrates strong current recognition with smooth planning degradation.}
\label{fig:qualitative}
\end{figure}

Figure~\ref{fig:qualitative} presents qualitative examples showing DARIL's recognition and planning capabilities. The model correctly identifies current actions while maintaining reasonable planning accuracy for short-term predictions, with graceful degradation over longer horizons.

\section{Discussion}

\subsection{Analysis: Why RL Underperformed}

Our analysis identifies key factors explaining RL's underperformance:

\textbf{Expert-Optimal Demonstrations}: CholecT50 contains expert-level data already near-optimal for evaluation metrics. RL exploration discovers valid alternatives that appear suboptimal under expert-similarity metrics.

\textbf{Evaluation Metric Alignment}: Test metrics directly reward expert-like behavior, giving IL fundamental advantages. This is common in medical domains where expert behavior defines gold standards.

\textbf{Limited Exploration Benefits}: Surgical domains have strong safety constraints limiting exploration benefits. While RL may discover novel approaches, these appear suboptimal for standard evaluation criteria.

\textbf{State-Action Representation Challenges}: Our RL implementations used frame embeddings as states and action triplets as discrete actions, with reward functions based on expert demonstration similarity. This design faced difficulties with comprehensive state representation and reward signal sparsity, potentially limiting learning effectiveness.

\textbf{Distribution Mismatch}: RL policies trained on different objective functions may produce valid but different behaviors that test metrics penalize due to expert demonstration alignment.

\subsection{Implications for Surgical AI}

Our findings have significant implications for surgical AI development:

\textbf{Method Selection}: In expert domains with high-quality demonstrations and aligned evaluation metrics, well-optimized IL may outperform sophisticated RL approaches. This challenges common assumptions about RL superiority in sequential decision making.

\textbf{Bootstrapping RL with IL}: Our findings suggest a promising approach for surgical AI: bootstrapping RL models with IL-learned basic skills, then using physics simulators or world models for safe exploration of new techniques, a strategy that aligns with emerging data-centric paradigms in the field~\cite{boels2025surgical}. This addresses the fundamental challenge that exploration and mistakes are crucial for improving surgical techniques, but must be tested in simulation rather than on patients.

\textbf{Safety Considerations}: IL approaches inherently stay closer to expert behavior, offering potential safety advantages in clinical deployment. RL exploration introduces uncertainty that may be undesirable in safety-critical surgical contexts.

\textbf{Clinical Translation}: Simpler IL models are easier to validate, interpret, and deploy in clinical settings compared to complex RL systems with learned reward functions.

\textbf{Cross-Dataset Generalization}: RL may prove advantageous when working across multiple datasets due to its ability to generalise more effectively, especially on non-expert trajectories. This suggests potential future work directions where RL's exploration capabilities could be beneficial for handling diverse surgical scenarios and skill levels.

\subsection{Limitations and Future Work}

Several limitations should be considered: (1) Single dataset evaluation on CholecT50 may not generalise to other surgical procedures. (2) Expert test data similar to training distributions may favor IL—results might differ with sub-expert or out-of-distribution scenarios. (3) Evaluation metrics directly reward expert-like behavior—alternative criteria focusing on patient outcomes might favor RL. (4) More sophisticated RL implementations with better state representations and reward design might outperform IL. (5) Overfitting concerns: Our models likely overfit to the limited number of videos and may not generalise well to test videos, indicating need for larger datasets and better simulators, whether world models or physics engines. (6) Lack of reward feedback: Our RL approaches suffered from insufficient reward signals from possible future states and lack of outcome data, limiting their learning effectiveness.

Future work should explore diverse surgical datasets, develop outcome-focused evaluation metrics, and investigate advanced RL techniques specifically designed for expert domains with comprehensive state-action-reward modeling.

\section{Conclusion}

This work provides crucial insights for surgical AI by demonstrating conditions where sophisticated RL approaches do not universally improve upon well-optimized IL. Our DARIL baseline consistently outperforms RL variants across planning horizons, with world model RL showing particularly poor performance (3.1\% vs 29.2\% at 10s).

The key insight is that expert domains with high-quality demonstrations may not benefit from RL exploration when evaluation metrics reward expert-like behavior. Distribution matching on expert-annotated test sets systematically favors IL over potentially valid RL policies that differ from training demonstrations.

Future surgical AI development should carefully consider domain characteristics, data quality, and evaluation alignment when choosing between IL and RL approaches. While IL excels at expert behavior cloning, RL's exploration capabilities may prove valuable in comprehensive evaluation frameworks capturing patient outcomes beyond expert similarity.

\bibliographystyle{splncs04}
\bibliography{references.bib} % e.g., references.bib

\end{document}